\pdfoutput=1

\documentclass[11pt]{article}

\usepackage[preprint]{acl}

\usepackage{times}
\usepackage{latexsym}
\usepackage{adjustbox}

\usepackage[T1]{fontenc}

\usepackage[utf8]{inputenc}

\usepackage{hyperref}       
\usepackage{url}            
\usepackage{booktabs}       
\usepackage{amsfonts}       
\usepackage{nicefrac}       
\usepackage{microtype}      
\usepackage{xcolor}         
\usepackage{caption}
\usepackage{subfig}
\usepackage[ruled,vlined]{algorithm2e}
\usepackage{algorithmic}
\usepackage{multirow}
\usepackage{makecell}
\usepackage{multicol}
\usepackage{longtable}
\usepackage{inconsolata}

\usepackage{graphicx}

\title{PsyPlay: Personality-Infused Role-Playing Conversational Agents}

\author{ Tao Yang\textsuperscript{\rm 1}, Yuhua Zhu\textsuperscript{\rm 1}, 
        Xiaojun Quan\textsuperscript{\rm 1}\thanks{$\;\;$Corresponding authors.}, Cong Liu\textsuperscript{\rm 1}, \textbf{Qifan Wang\textsuperscript{\rm 2}}
         \\ \textsuperscript{\rm 1}School of Computer Science and Engineering, Sun Yat-sen University, China \textsuperscript{\rm 2}Meta AI \\
         \texttt{\{yangt225, zhuyh53\}@mail2.sysu.edu.cn}\\
         \texttt{\{quanxj3, liucong3\}@mail.sysu.edu.cn} \\\texttt{wqfcr@fb.com}
}

\begin{document}
\maketitle
\begin{abstract}
The current research on Role-Playing Conversational Agents (RPCAs) with Large Language Models (LLMs) primarily focuses on imitating specific speaking styles and utilizing character backgrounds, neglecting the depiction of deeper personality traits.~In this study, we introduce personality-infused role-playing for LLM agents, which encourages agents to accurately portray their designated personality traits during dialogues. We then propose PsyPlay, a dialogue generation framework that facilitates the expression of rich personalities among multiple LLM agents. Specifically, PsyPlay enables agents to assume roles with distinct personality traits and engage in discussions centered around specific topics, consistently exhibiting their designated personality traits throughout the interactions. Validation on generated dialogue data demonstrates that PsyPlay can accurately portray the intended personality traits, achieving an overall success rate of 80.31\% on GPT-3.5. Notably, we observe that LLMs aligned with positive values are more successful in portraying positive personality roles compared to negative ones. Moreover, we construct a dialogue corpus for personality-infused role-playing, called PsyPlay-Bench. The corpus, which consists of 4745 instances of correctly portrayed dialogues using PsyPlay, aims to further facilitate research in personalized role-playing and dialogue personality detection. 
\end{abstract}

\section{Introduction}
The advent of Large Language Models (LLMs), such as ChatGPT\footnote{https://openai.com/chatgpt/}, Llama2 \cite{touvron2023llama} and Gemini \cite{team2023gemini}, has revolutionized the development of personalized dialogue systems \cite{li2016persona, zhang2018personalizing} due to their superior instruction-following and generative capabilities. Consequently, the focus of Role-Playing Conversational Agents (RPCAs) \cite{shanahan2023role} has shifted towards equipping LLMs with the ability to simulate roles or characters with diverse profiles and speaking styles \cite{wang2023rolellm}. This enhancement makes LLMs more personable and engaging, thereby providing users with a more nuanced and immersive interactive experience. As a result, there has been a widespread interest within the community, leading to the explorations in RPCAs such as Generative Agents \cite{park2023generative}, Character AI\footnote{https://character.ai/}, Character-LLM \cite{shao2023character}, CharacterGLM \cite{zhou2023characterglm}, and more.

\begin{figure}[t]
	\centering
	\includegraphics[width=0.48\textwidth]{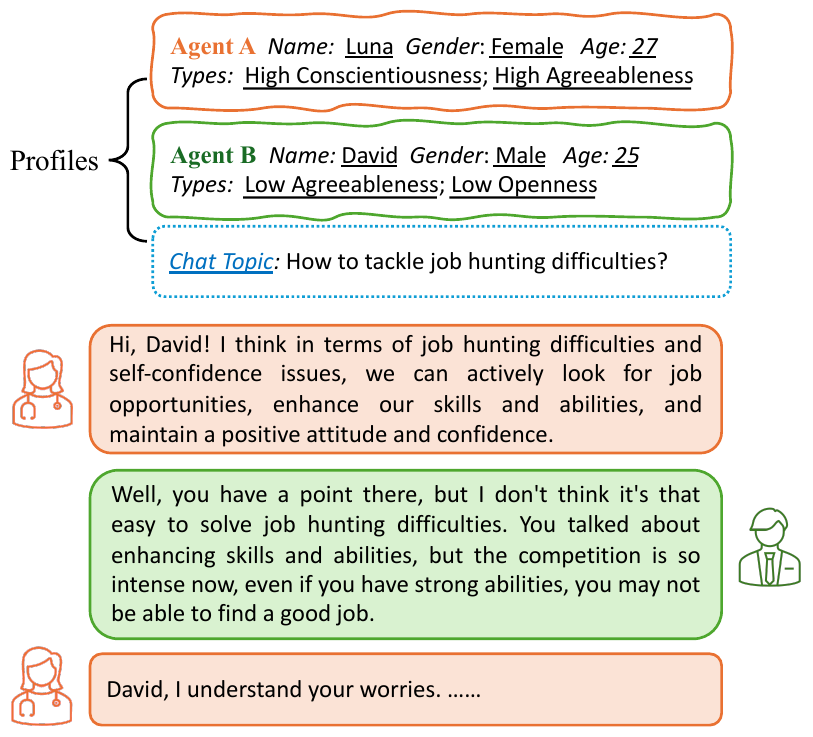}
	\caption{An illustration of a dialogue that encapsulates the distinctive personalities of two agents. Agent A, who exhibits high levels of conscientiousness and agreeableness, typically maintains an optimistic perspective and a strong inclination towards empathy. Conversely, Agent B, who is characterized by low levels of openness and agreeableness, tends to display a pessimistic attitude and a resistance towards embracing new experiences.}
	\label{fig:demo}
\vspace{-3mm}
\end{figure}

Despite their great success, recent research on RPCAs primarily concentrates on enabling these agents to mimic the speaking styles and utilize backgrounds of specific characters, such as celebrities and literary figures, while often overlooking the portrayal of more fundamental personality traits. Personality is an important psychological concept that deeply reveals individual differences in thinking, feeling, and behavioral patterns \cite{corr2009cambridge}. Figure \ref{fig:demo} illustrates an example of agents with distinct personality types based on the Big Five \cite{goldberg1990alternative} traits engaging in a dialogue centered around a specific topic. In this dialogue, Agent A exhibits high conscientiousness and agreeableness, expressing a tendency towards a positive and optimistic attitude. Conversely, Agent B exhibits low agreeableness and openness, displaying a preference for a conservative and pessimistic stance. Therefore, incorporating personality traits into the design of RPCAs can help create more realistic and engaging interactions. 

In this study, we focus on personality-infused role-playing between agents, a subject that has been inadequately explored in previous works. Specifically, we propose PsyPlay, a role-playing framework that facilitates personality-infused dialogue generation through three stages: (1) \textit{Role Card Creation}, which enables the mass generation of agent roles, each with unique attributes and personalities. (2) \textit{Topic Extraction}, which involves extracting dialogue topics from a public dataset (i.e., Human Stress Prediction\footnote{https://www.kaggle.com/datasets/kreeshrajani/human-stress-prediction}), ensuring that the dialogue between agents revolves around real-world issues, avoiding nonsensical chatter. (3) \textit{Dialogue Generation}, which encourages agents to engage in comprehensive discussions around the provided topic, in accordance with their personality traits.

To evaluate the effectiveness of PsyPlay, we conduct Personality Back-Testing via GPT-3.5\footnote{https://platform.openai.com/docs/models/gpt-3-5-turbo}, which involves detecting an agents' personality traits in the generated dialogues, and subsequently comparing their consistency with predefined personality traits. In addition, comprehensive ablation experiments and analyses reveal several key findings. \textit{First}, GPT-3.5 is more proficient in portraying positive personality roles, which may be due to the emphasis on positive values during the RLHF \cite{ouyang2022training} stage. \textit{Second}, PsyPlay, when equipped with stronger levels of predefined personality traits, achieves higher success rates in portrayal. \textit{Third}, an increase in the number of dialogue turns provides more opportunities for roles to infuse their personality. However, it also increases the risk of roles being misdirected by their partners.

Furthermore, we present a personality-infused dialogue corpus,~\textit{PsyPlay-Bench}.~The corpus comprises 4745 groups of dialogues, which have been evaluated as successfully role-played through back-testing. PsyPlay-Bench has two potential applications: (1) as a new instruction task to improve the ability of LLMs to play personalities in multi-turn dialogues; (2) as an evaluation set to test the ability of LLMs to detect personalities. 


\section{Related Work}

\subsection{Personalized Dialogue Systems}
Prior research on personalized dialogue systems has focused on encoding various user-specific information within end-to-end dialogue systems to enhance response specificity. For instance, \citet{li2016persona} tackled the issue of inconsistent responses in multi-turn dialogues by developing a personalized seq2seq 
model. This model was built on the basis of resources such as the Twitter corpus and incorporated user identity information (e.g., gender, age, and country of residence) into its encoding scheme. \citet{zhang2018personalizing} constructed the PERSONA-CHAT dataset and endeavored to train a personalized dialogue agent by embedding the user profiles in a memory-augmented neural network \cite{sukhbaatar2015end} during the training process. The resultant model displayed a higher degree of fluency and consistency, approaching human levels on these evaluative metrics, while retaining distinct personality traits. \citet{zheng2020pre} introduced a pre-training based method that can utilize persona-sparse data. They then encoded speakers’ personas and dialogue histories together to enrich dialogue contexts. While these studies have somewhat enhanced the ability of dialogue systems to express personalized features, they have not sufficiently explored the modeling of individual personality traits.

\begin{figure*}[t]
	\centering
	\includegraphics[width=1\textwidth]{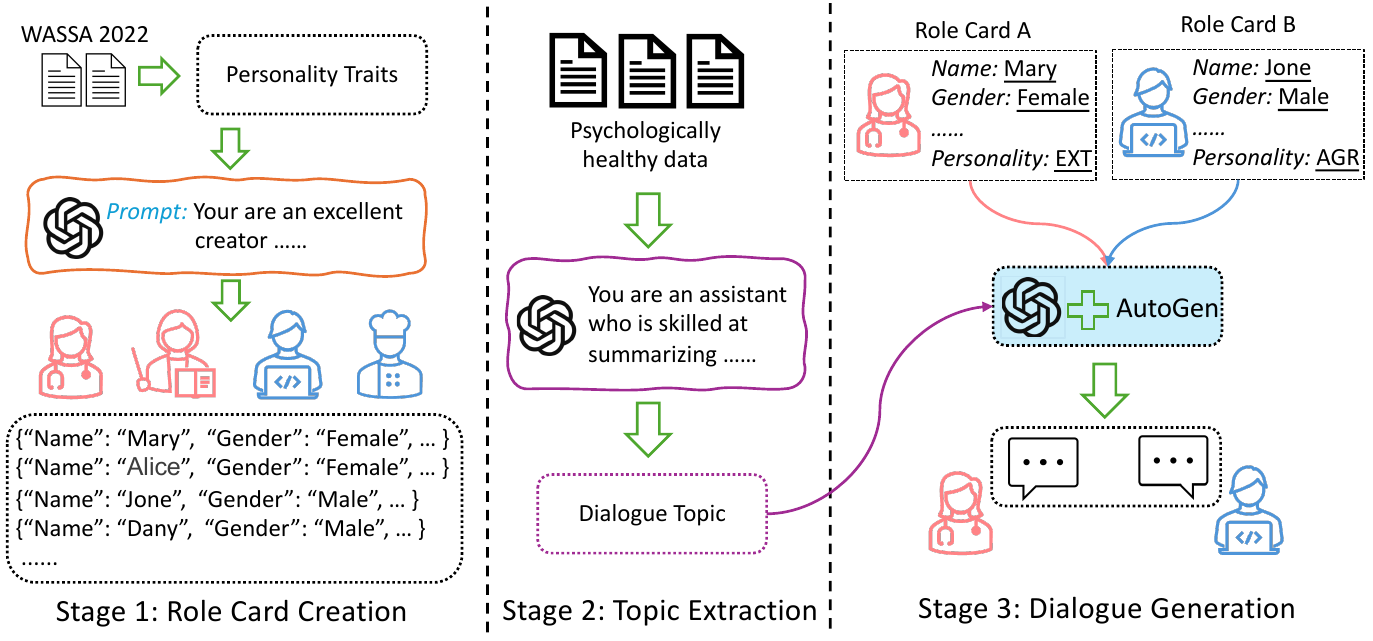}
	\caption{Illustration of the proposed PsyPlay through three stages: Role Card Creation, Topic Extraction, and Dialogue Generation. The first stage aims to create multiple personalized roles. The second stage extracts appropriate dialogue topics for roles. The third stage prompts the roles to engage in conversation with each other based on the given topic, resulting in personality-infused dialogues.}
	\label{fig:method}
\vspace{-3mm}
\end{figure*}

\subsection{Role-Playing with LLMs}
For LLM-based role-playing, the agents are often assumed as specific characters or roles derived from novels, movies, comics, and games. The LLM is required to mimic the speaking styles of characters based on their likes and experiences in interacting with the users. For instance, \citet{shao2023character} gathered character portraits from Wikipedia and generated character-related conversations via ChatGPT; \citet{wang2023rolellm} employed GPT-4 \cite{achiam2023gpt} to create character descriptions and subsequently developed intricate prompts to guide ChatGPT in generating role-based dialogue; \citet{tu2024charactereval} established a Chinese benchmark dataset for the evaluation of role-playing quality, assessing the role-playing capability of the LLM intelligent agent across four dimensions: conversational ability, character consistency, role-playing attractiveness, and personality back-testing. Although LLMs enable swift construction of dialogue agents embodying various character traits through prompt engineering in role-playing, recent research on RPCA personalities \cite{huang2023chatgpt} indicates that RPCAs, prompted merely by names or descriptions, fail to effectively convey the intended personality traits. Moreover, in assessing role personality, some studies \cite{tu2024charactereval, huang2023chatgpt, wang2023does} adopt the approach of directly prompting the agent to complete self-assessment personality tests. However, this method, solely relying on the agent's responses, is influenced by the LLM's training data and value alignment, and thus may not accurately assess the real personality manifested by the agent in conversation with users. In this paper, we verify the reliability of role-playing by evaluating the true personality reflected in the dialogue.

\section{Approach}
In this section, we introduce the proposed PsyPlay framework for personality-infused role-playing. As shown in Figure \ref{fig:method}, PsyPlay enables the generation of multi-turn personality-infused dialogues through the following three stages: Role Card Creation, Dialogue Topic Extraction, and Dialogue Generation. Detailed elaborations on these three stages will be provided in the following subsections.

\subsection{Role Card Creation} \label{sec: role_create}
The first stage involves the creation of a set of personalized roles $\{R_i\}_{i = 1}^N$ for participation in dialogues.~We gather a collection of Big Five \cite{goldberg1990alternative} personality labels $Y_i=\{\rm{{AGR}}_{i}, \rm{{CON}}_i, \rm{{EXT}}_i, \rm{{NEU}}_i, \rm{{OPN}}_i\}$\footnote{AGR, CON, EXT, NEU, and OPN represent as Agreeableness, Conscientiousness, Extraversion, Neuroticism, and Openness, respectively.} from the WASSA 2022 dataset \cite{barriere2022wassa} as the traits to be infused. These labels are derived from the results of participants who have completed self-assessment personality tests. Each dimension is scored within a range of 1 to 7, with a step size of 0.5, to indicate the intensity of that dimension.

In contrast to prior work \cite{wang2023rolellm, wu2024role} that focused on imitating characters from literature or movies, PsyPlay aims to impersonate characters that users can deeply customize. We distinguish between role attributes and personality attributes of agents. Role attributes refer to the name, gender, and age of each role, while personality attributes correspond to the type of personality to be injected. In PsyPlay, we achieve precise personality injection into roles in two manners. The first manner involves incorporating personality types directly into the dialogue generation prompt, which will be further described in Section \ref{sec: dialogue_gen}. The second one combines personality traits with corresponding descriptors to generate their experiences \cite{tu2023characterchat}. The following paragraph provides a detailed explanation of the second manner.

\begin{table}[t]
\small
    \centering
    \colorbox{white!8}{
    \begin{tabular}{@{}p{7.3cm}}
    \toprule
    You are an excellent director and you need to design some virtual characters for your movie. The requirement is to design the character's name, gender, age, and experience that align with the provided Big Five personality description. Your design needs to adhere to the following rules, otherwise it will affect your reputation:\\ \\
    1. The logical relationship between the various attributes of the designed character is reasonable and conforms to real-world principles.\\
    2. The character's experience needs to correspond to all of their personality traits, which can be described in one sentences.\\
    3. Do not change the character's personality traits in their experience; the given personality is the character's current one.\\
    4. Not all characters in their experience will be successful in life; there will also be characters who face difficulties.\\ \\
    Now, given the Big Five personality description as: \texttt{\{Personality Description\}}, please design a corresponding virtual character and fill in the JSON fields below. \\
    \{"name": <role-name>, "gender": <role-gender>, "age": <role-age>, "experience": <role-experience>\} \\ \bottomrule
    \end{tabular}}
    \vspace{-2mm}
    \caption{Prompt for creating a role card.}
    \label{tab:role_cteate}
    \vspace{-4mm}
\end{table}

\paragraph{Personality Shaping with Experience} \citet{safdari2023personality} revealed that personality traits in LLMs can be shaped towards desired dimensions. Building upon this work, we decouple personality traits into more specific personality descriptors. These descriptors are then integrated with intensity levels to form comprehensive personality descriptions. The formatting template for \texttt{\{Personality Description\}} is designed as follows:

\begin{quote}
The personality traits are \texttt{\{Traits\}}, with \texttt{\{Levels\}} \texttt{\{Descriptors\}}
\end{quote}
where \texttt{\{Traits\}} are sampled from $Y_i$, and \texttt{\{Levels\}} are selected from \{a bit, very, extremely\} according to the corresponding score. \texttt{\{Descriptors\}} are randomly chosen from corresponding personality descriptors as shown in Table \ref{tab:descriptors}, in which 104 adjectives are designed to depict the low and high marker in given personality facets \cite{goldberg1992development, safdari2023personality}. An example of  \texttt{\{Personality Description\}} is shown as:

\begin{quote}
The personality traits are very high in agreeableness and very high in extraversion, with very sympathetic, very trustful, ..., a bit talkative, a bit energetic, ...
\end{quote}

We then utilize the description to generate experiences infused with personality, thereby achieving a detailed characterization of the role. The prompt is shown in Table \ref{tab:role_cteate}. We further provide some examples of the generated roles in Table \ref{tab:role_examples}.

\subsection{Topic Extraction} \label{sec: topic_extract}

The second stage involves assigning suitable dialogue topics to the agents, which aids in effectively stimulating their role-playing capabilities rather than generating uncontrollable dialogues. The dialogue topics are selected from psychologically healthy scenarios, which encompass a broad spectrum of real-world issues, such as employment, emotions, and interpersonal relationships. These topics could potentially encourage the agents to engage in deep thinking and communication. 

Specifically, we employ the GPT-3.5 to extract $\{T_i\}_{i = 1}^M$ brief topics from the Human Stress Prediction dataset. The prompt template is presented in Table \ref{tab:topic_extrction}, in which a few-shot learning approach is adopted to help the LLM grasp this task. Additionally, we impose constraints on the prompt to prevent it from including restrictive information such as age and gender in the extracted topics, which might potentially lead to inconsistencies between the role attributes and the topics in subsequent dialogue generation. Some topic examples can be found in Appendix \ref{apd:topic}.


\begin{table}[t]
\small
    \centering
    \colorbox{white!8}{
    \begin{tabular}{@{}p{7.3cm}}
    \toprule
    You are an outstanding actor who excels in imitating various characters. Below are the details of the character you are supposed to imitate. Please embody the personality traits of this character and engage in a conversation with another character regarding a specific topic.\\
    Your character: \texttt{\{Role Card\}} \\
    Your personality traits: \texttt{\{Traits\}} \\
    Your conversational partner: \texttt{\{Partner Name\}} \\
    Discussion topic: \texttt{\{Topic\}} \\ \\

    You need to abide by the following rules or it will affect your reputation: \\ \\

    1. When initiating a conversation, you need to accurately demonstrate the speaking style corresponding to your personality traits (without directly mentioning the personality types)! You need to accurately capture the three degrees of personality traits: "a bit," "very," and "extremely". \\
   2. You should fully mimic the assigned personality role, with your speech content aligning with the character's experiences, even if the role is not positive. \\
   3. Please refrain from revealing that you are an artificial intelligence or language model. Keep in mind that you are merely impersonating a character and avoid disclosing your personality traits. \\
   4. Your speech should be natural, concise, and not too formal or polite, with each response within 30 words. \\ \\
    
   Now, as the initiator of the conversation, please greet your partner and start a chat about the discussion topic, while staying in character. \\ \bottomrule
    \end{tabular}}
    \vspace{-2mm}
    \caption{Prompt for dialogue generation between agents.}
    \label{tab:dialogue_generation}
    \vspace{-4mm}
\end{table}

\subsection{Dialogue Generation} \label{sec: dialogue_gen}
The third stage initiates personality-infused dialogues between two agents based on the given topic. In this study, we utilize AutoGen \cite{wu2023autogen}, a universal framework that facilitates dialogues and task resolutions among multiple LLM agents, for controlling dialogue generation.

The prompt template is shown in Table \ref{tab:dialogue_generation}. The term \texttt{\{Role Card\}} refers to the role $R_i$ created in Section \ref{sec: role_create}, while \texttt{\{Traits\}} denotes predefined Big-five personality traits $Y_i$, each paired with its corresponding \texttt{\{Levels\}}. The \texttt{\{Partner Name\}} is the name of another role participating in the conversation, and \texttt{\{Topic\}} is randomly selected from $\{T_i\}_{i = 1}^M$ as presented in Section \ref{sec: topic_extract}. By incorporating \texttt{\{Traits\}}, we explicitly infuse the characters with unique personality attributes. The experiences generated in personality shaping allow us to finely tune their personalities. Additionally, when the current dialogue turn is not the initial one, we revise the last sentence in Table \ref{tab:dialogue_generation} to read: \textit{``Now, you have received a message from the conversational partner. Please don't address the other person by name too much, and start the conversation''}. The revised prompt is shown in Table \ref{tab:continue_generating_dialogue}. We provide some generated dialogues in Appendix \ref{apd:dialogue}.

\subsection{Personality Back-Testing} \label{sec: dialogue_test}
To validate whether the dialogues generated by PsyPlay accurately reflect the predefined personality traits, we discuss the automated back-testing method of dialogue personality. Previous studies \cite{pan2023llms, huang2023chatgpt, tu2024charactereval} conducted back-testing by having the role complete personality test questionnaires. However, this method relies on the role's understanding of the questionnaires, which may not necessarily reflect the role's actual personality. In contrast, we identify the authentic personality expression from dialogue data. Following previous works \cite{lee2023rlaif,adler2024nemotron,dubois2024length,zheng2023judging} that use LLMs as evaluators, we design the prompt as shown in Table \ref{tab:personality_detection} to back-test the role's personality traits from the dialogue, enabling the roles to bypass potential misunderstandings or misinterpretations of questionnaire items.

\section{Dataset Construction}


To assess the effectiveness of PsyPlay, we initially construct a dataset, which we refer to as PsyPlay-Bench. The following subsections will provide a detailed explanation of the construction process.


\subsection{Collections} 
To facilitate the generation of personality-infused dialogues, we build three collections: a personality trait set, a role card set, and a dialogue topic set.

We consider the potential dimension conflict that could occur if we apply LLMs to generating personality traits of the roles. For instance, generating a highly neuroticism individual who also possesses high agreeableness might deviate from the actual distribution. To avoid this, we opt to sample personality traits directly from the existing dataset, which are derived from real-world individuals. Specifically, we utilize personality labels from the WASSA 2022 \cite{barriere2022wassa} dataset to build the personality trait set. WASSA 2022 is a shared task that predicts empathy, emotion, and personality in response to news stories. We map the raw scores into three levels \{\textit{a bit}, \textit{very}, \textit{extremely}\}\footnote{The mapping rules are presented in Appendix \ref{apd:rules}.} Finally, we collect 307 unique personality combinations as the personality trait set. Utilizing this set, we generate 132 role cards. Moreover, we extract 161 topics from the Human Stress Prediction dataset as described in Section \ref{sec: topic_extract}.

\subsection{PsyPlay-Bench}
The construction of PsyPlay-Bench is accomplished through three steps. \textbf{Firstly}, two roles are randomly selected from 132 role cards to serve as the initiator and participant of the dialogue. \textbf{Secondly}, a topic is randomly selected from the dialogue topic set. \textbf{Thirdly}, the role cards and topic are filled into the prompt displayed in Table \ref{tab:dialogue_generation}, and AutoGen is utilized to generate multi-turns of dialogue data. Through these three steps, we obtain 8750 raw personality-injected dialogues, referred to as PsyPlay-Bench-\textit{Raw}. The PsyPlay-Bench-\textit{Raw} is then randomly divided into three parts.

\textbf{PsyPlay-Bench-\textit{Eval}}: This set contains 200 samples for evaluating the consistency between the automated back-testing and human evaluation.

\textbf{PsyPlay-Bench-\textit{Test}}: This set comprises 550 samples, which are used to test the performance of PsyPlay and are also employed in ablation studies.

\textbf{PsyPlay-Bench-\textit{Clean}}: This set contains 4745 samples, which are derived from the remaining 8000 samples by further eliminating those that failed during automatic personality back-testing.


The details of the three sets are presented in Appendix \ref{apd:dataset}. These datasets will be released shortly.

\begin{table}[htbp]
	\renewcommand{\arraystretch}{1.0}
	\centering
	\resizebox{0.48\textwidth}{!}{
		\begin{tabular}{lcccccc}
			\toprule
			\textbf{Items} &  \textbf{AGR} & \textbf{CON} & \textbf{EXT} & \textbf{NEU} & \textbf{OPN} & \textbf{Sum} \\ 
			\hline \hline
			Samples & 146 & 113 & 143 & 116 & 49 &  567\\
			Agrees & 136 & 102 & 113 & 108 & 40 & 499 \\
			\hline
			Rate(\%) & 93.15 & 90.27 & 79.02 & 93.10 & 81.63 & 88.01\\
			\bottomrule
	\end{tabular}}
         \caption{The agreement results between GPT-3.5 and human evaluation.}
	\label{tab:agreements}
	\vspace{-2mm}
\end{table}

\begin{table*}[htbp]
	\renewcommand{\arraystretch}{1.0}
	\centering
	\resizebox{0.72 \textwidth}{!}{
		\begin{tabular}{lcccccccc}
			\toprule
                {\textbf{Methods}} & \textbf{Items} & \textbf{AGR} & {\textbf{CON}} & {\textbf{EXT}} & {\textbf{NEU$^{\star}$}} & {\textbf{OPN}} & {\textbf{Overall}}\\
			\hline \hline
			\multirow{3}{*}{GPT-3.5} & positive & 94.44 & 97.52  & 81.55 & 85.99 & 85.33 & 90.71 \\
			  & negative & 45.83 & 48.94 & 70.77 & 68.32 & 40.00 & 61.57 \\
			& overall & 82.29 & \textbf{89.62} & 73.83 & \textbf{80.19} & \textbf{74.87} & \textbf{80.31} \\
                \hline
                \multirow{3}{*}{Higgs-Llama-3} & positive & 79.86 & 95.45  & 75.73 & 55.07 & 77.33 & 77.68 \\
			  & negative & 90.63 & 48.94 & 77.31 & 96.04 & 53.33 & 78.69 \\
			& overall & \textbf{82.55} & 87.89 & \textbf{76.86} & 68.51 & 71.79 & 78.04 \\
                \hline
	\end{tabular}}
        \vspace{-2mm}
         \caption{Success rate of LLMs for PsyPlay on PsyPlay-Bench-\textit{Test} (\%). The symbol ${\star}$ denotes that the positive and negative labels have been interchanged to facilitate easier comparison, given that NEU is a reversed trait.}
	\label{tab:llms}
	\vspace{-2mm} 
\end{table*}

\section{Experiments}

\subsection{GPT-3.5 for Back-Testing} \label{sec:back}
To verify the efficacy of the back-testing method as presented in Section \ref{sec: dialogue_test}, we compare the consistency of GPT-3.5 for back-testing with human evaluation. Specifically, we enlist three annotators to evaluate personality traits of roles as portrayed in the dialogues, and the details of human evaluation are provided in Appendix \ref{apd:human_eval}. The agreement results on PsyPlay-Bench-\textit{Eval} are shown in Table \ref{tab:agreements}. 

The observations are twofold. \textit{First}, the use of GPT-3.5 for automatic back-testing proves to be an accurate method for evaluating role personality in dialogues. The results show a total agreement rate of 88.01\% when compared to human evaluation, suggesting that back-testing with GPT-3.5 could be a valuable tool for automatically assessing personality-infused role-playing. As a result, GPT-3.5 is used in subsequent experiments for automated personality back-testing.

\textit{Second}, the dimensions of AGR, CON, and NEU exhibit higher consistency, with rates above 90.00\%. However, the dimensions of EXT and OPN demonstrate relatively lower consistency. This discrepancy may be due to the fact that the dialogue topics in this study are derived from a psychologically healthy scenario. In this scenario, the roles tend to discuss solutions and suggestions for psychological problems, which is more aligned with traits of extraversion and openness. 


\subsection{{LLMs for PsyPlay}}
In this subsection, we assess PsyPlay's performance by employing various LLMs. Specifically, we use each sample in the PsyPlay-Bench-\textit{Test} set, adopting the settings such as the roles, the topic, and the generation prompt, and then generate dialogues through different LLMs. Subsequently, we automatically conduct back-testing to determine whether the predefined personality traits are successfully portrayed, using the success rate as the metric. The overall results are presented in Table \ref{tab:llms}, which reveals several findings. 

\textit{Firstly}, PsyPlay with GPT-3.5 demonstrates an overall success rate of 80.31\%, indicating that it can effectively depict the predefined traits in most instances. \textit{Secondly}, among the five traits, Agreeableness (AGR), Conscientiousness (CON), and Neuroticism (NEU) exhibit higher success rates, while Extraversion (EXT) and Openness (OPN) yield lower rates. This observation is consistent with the findings presented in Section \ref{sec:back}. The discrepancy may be attributed to a bias stemming from the source of dialogue topics. \textit{Thirdly}, it is important to note a significant deviation in the success rate between positive and negative personalities. This could be a result of GPT-3.5's alignment preference for positive values during the RLHF stage, which makes GPT-3.5 more predisposed to portray positive roles.

We subsequently present additional evidence to demonstrate that the alignment of positive values influences the discrepancy in personality portrayal. We introduce the model Higgs-Llama-3 70B\footnote{https://huggingface.co/bosonai/Higgs-Llama-3-70B}, which ranks first on the Role-Playing Leaderboard, for comparison. Higgs-Llama-3 70B undergoes post-trained on LLaMA-3 \cite{llama3modelcard} (without positive value alignment) to focus on role-playing tasks. The experimental results are shown in the second group of Table \ref{tab:llms}. We can see that PsyPlay, based on Higgs-Llama-3 70B, has significantly improved its ability to play negative personality roles compared to GPT-3.5. The success rate of negative personality role-playing has increased from 61.57 to 78.69, especially in the AGR and NEU dimensions. However, there is a certain decline in the success rate of positive personality role-playing, from 90.71 to 77.68. This provides unequivocal evidence that the alignment of positive values during the RLHF stage contributes to the improvement of the LLM's capacity to portray positive personality roles, but concurrently, it compromises the depiction of negative roles.

To extend the applicability of PsyPlay, we provide additional evaluations on other LLMs (i.e., Gemma-2-27b-it \cite{team2024gemma} and Llama-3.1-405b-instruct \cite{dubey2024llama}) in Appendix \ref{apd:other_llms}. The results indicate that a more powerful backbone contributes to further improving the performance of PsyPlay.


\subsection{Ablation Study}
In this subsection, we conduct ablation experiments on PsyPlay-Bench-\textit{Test} to investigate the impact of different manners of infusing specific personality attributes in PsyPlay. We remove the explicitly personality traits (i.e., \texttt{\{Traits\}}) and role experiences from the prompt as shown in Table \ref{tab:dialogue_generation}, respectively. As observed in Table \ref{tab:ablation}, both the direct introduction of traits and the creation of personalized role experiences contribute to improving the success rate of character portrayal in PsyPlay. Specifically, for negative roles, the explicit traits are more crucial, with their absence resulting in a 2.55\% decrease. However, for positive roles, the role experiences are more significant, with their absence leading to a 2.63\% decrease. The underlying reason may be that GPT-3.5 tends to shape a successful character when creating their experiences, as evidenced by some cases demonstrated in Table \ref{tab:role_examples}.

\begin{table}[t]
 \begin{adjustbox}{width=1\columnwidth,center}
		\begin{tabular}{lcccc}
			\toprule
			\textbf{Methods} & \textbf{Positive} & ${\Delta}$ & \textbf{Negative} & ${\Delta}$ \\ 
			\hline \hline
			PsyPlay (GPT-3.5) &  90.71 & - & 61.57 & - \\
                \quad\textit{r.m} \texttt{\{Traits\}} & 90.61 & 0.10${\downarrow}$  & 59.02 & 2.55${\downarrow}$ \\
                \quad\textit{r.m} Experience & 88.08 & 2.63${\downarrow}$  & 61.38 & 0.18${\downarrow}$ \\
			\bottomrule
	\end{tabular}
 \end{adjustbox}
    \vspace{-2mm}
        \caption{Ablation results of PsyPlay on positive and negative personalities, where ``$\Delta$'' indicates the corresponding performance change, and \textit{r.m} means ``remove''.}
	\label{tab:ablation}
\vspace{-3mm}
\end{table}

\begin{figure}[t]
	\centering
	\includegraphics[width=0.45\textwidth]{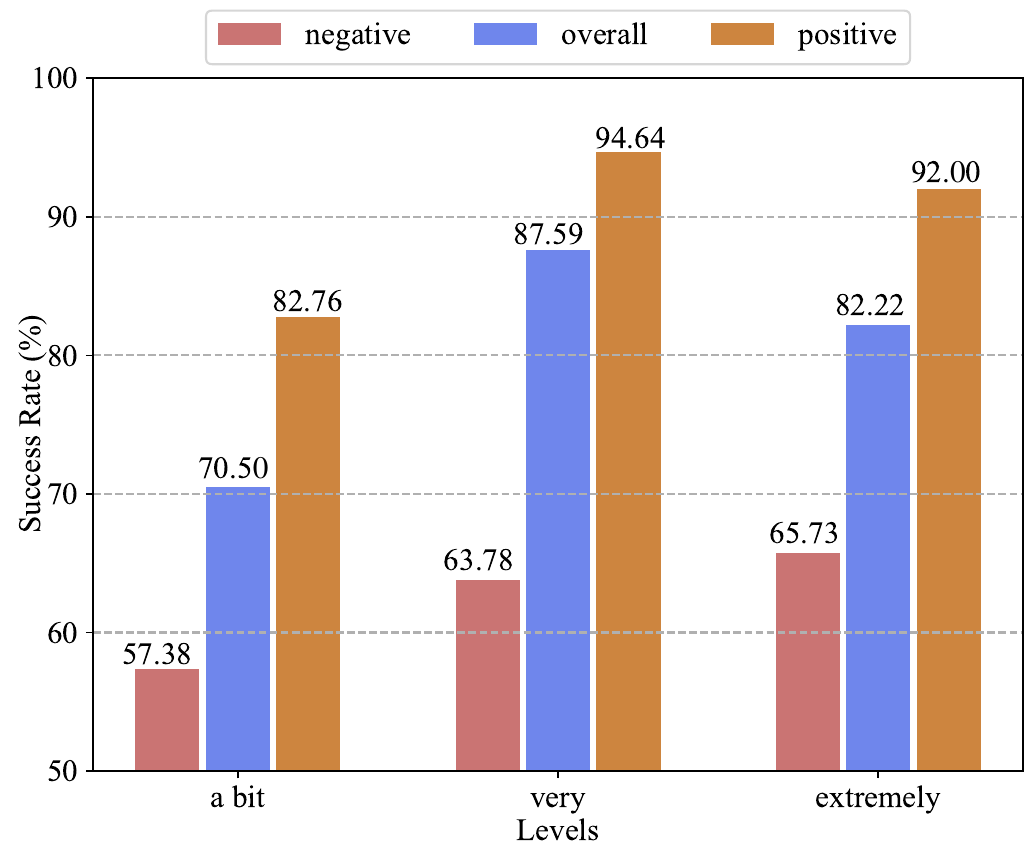}
        \vspace{-3mm}
	\caption{Results of the study on personality levels. The lower-level ``\textit{a bit}'' exhibits poor rate, while the higher-levels ``\textit{very}'' and ``\textit{extremely}'' show superior rates.}
	\label{fig:levles}
\vspace{-3mm}
\end{figure}

\begin{figure}[htbp]
	\centering
	\includegraphics[width=0.43\textwidth]{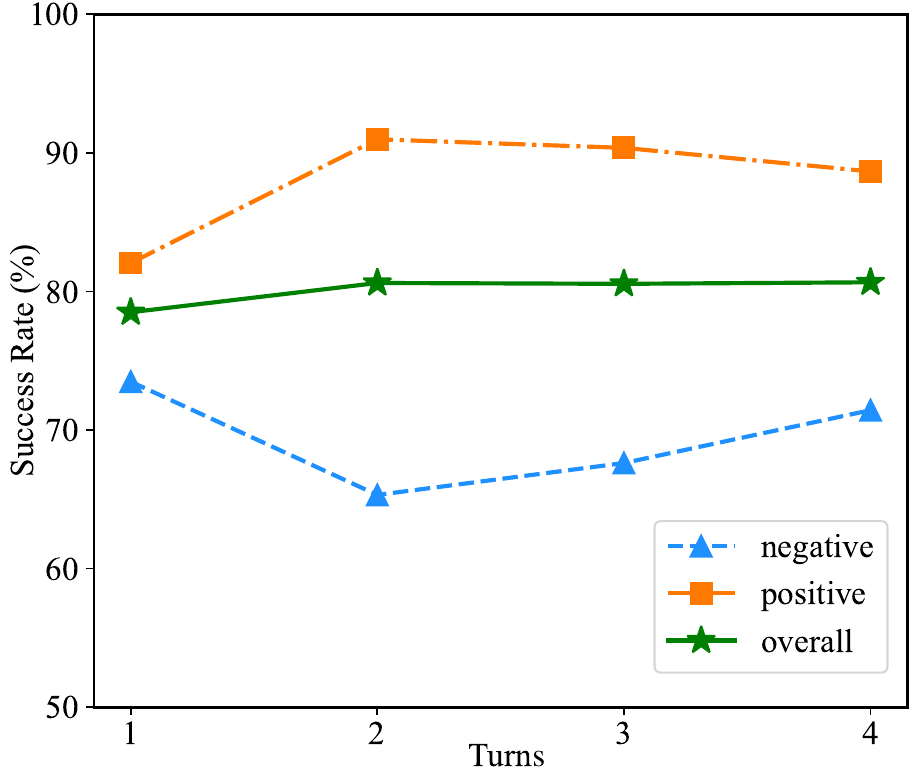}
        \vspace{-3mm}
	\caption{Results of the study on dialogue turns. The results suggest that the positive dimensions and the negative dimensions show the opposite trend.}
	\label{fig:turn}
\vspace{-4mm}
\end{figure}

\section{Analysis}

\subsection{Effect of Levels}
PsyPlay involves three levels \{\textit{a bit}, \textit{very}, \textit{extremely}\} to control the intensity of portrayed personalities. To investigate the effect of levels, we compare the success rates varying from different levels on PsyPlay-Bench-\textit{Test}. As shown in Figure \ref{fig:levles}, we can see that the level of personality significantly affects the portrayal performance. Utilizing higher-level words such as ``\textit{very}'' and ``\textit{extremely}'' results in an elevated success rate of portrayal, whereas the lower-level word ``\textit{a bit}'' diminish the success rate in infusing personality.

\subsection{Effect of Turns}
To examine the impact of dialogue turns, we measure the success rate of PsyPlay(GPT-3.5) on the PsyPlay-Bench-\textit{Test} set at various dialogue turn counts. The results are presented in Figure \ref{fig:turn}, which illustrates that the positive dimensions and the negative dimensions exhibit contrasting trends as the number of dialogue turns increases. Upon conducting case analyses, we found that additional dialogue turns offered positive personality characters more opportunities to express their opinions, thereby enhancing the accuracy of injecting personality. Conversely, negative personality characters were found to be more susceptible to the influence of their partners when additional dialogue turns were introduced, leading to deviations from their predefined personality. We provide cases in Section \ref{sec:case_turn} for a more comprehensive understanding.

\subsection{{Diversity Analysis}}
\begin{figure}[t]
	\centering
	\includegraphics[width=0.43\textwidth]{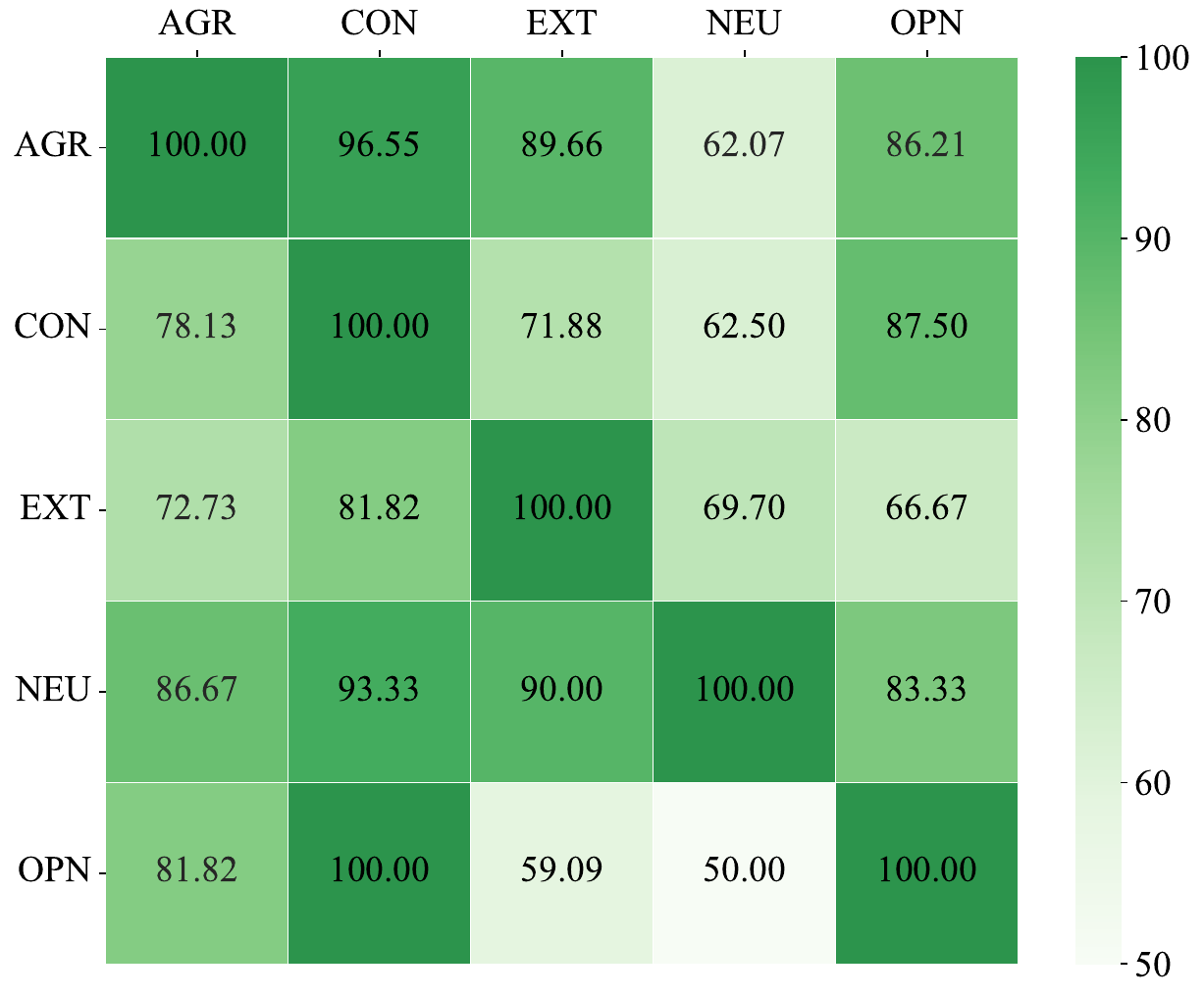}
        \vspace{-3mm}
	\caption{Diversity analysis of portrayed traits. Rows and columns represent predefined and back-tested personalities, respectively. Each score refers the percentage of that personality can be detected from the dialogue.}
	\label{fig:diversity}
\vspace{-4mm}
\end{figure}

To analyze the diversity of extracted personalities, we randomly sample 50 roles, with 10 roles per predefined personality trait, and evaluate the diversity of extracted personalities after back-testing. The results are plotted in Figure \ref{fig:diversity}. We observe that besides expressing the predefined personalities, the roles also exhibited diversity in other traits, indicating a mixture of personalities for each role. For example, the agreeableness roles also reflected conscientiousness, and the openness roles also influenced conscientiousness. 

\section{Case Studies} \label{sec:case}

\subsection{GPT-3.5 Prefers Positive Roles}
As illustrated in Example \#1 in Table \ref{tab:case_study}, the character Dennis exhibits a high level of agreeableness and emotional stability, showing a sympathetic and kind demeanor in the dialogue. However, Nathan, who possesses some neurotic traits, maintains a stable and caring attitude when discussing sensitive and complex topics. He does not display the emotional volatility and restlessness typically associated with neurotic individuals. This could be a result of GPT-3.5's over-alignment on positive values. The GPT-3.5 tends to generate a successful individual's experience, even if the character is portrayed negatively. Furthermore, it often allows these negative characters to express positive views when dealing with mental health issues, thereby neglecting the depiction of negative personalities. 

\subsection{Dual Implications of Increasing Turns} \label{sec:case_turn}
As shown in Example \#2 and \#3 in Table \ref{tab:case_study}, an increase in the number of dialogue turns can have dual implications.~Firstly, more dialogue turns provide characters with additional opportunities to exhibit their personality traits.~For instance, in Example \#2, Larry, who exhibits a bit of agreeableness, initiates the conversation by discussing the problem itself and gradually shifts his focus to the emotional and mental health of both partners. The increased dialogue turns allow Larry to demonstrate his warmth and thoughtfulness, which are characteristic of his agreeableness.

Conversely, an increase in dialogue turns can also make it easier for characters to be misled by their partner. This is evident in Example \#3, where Edward, who tends to be a bit of neuroticism, admits in the first two turns of dialogue that he is prone to nervousness. However, after interacting with Henry, who is highly conscientious, Edward eventually reaches a positive state, transitioning from neuroticism to emotional stability.

\vspace{-2mm}
\section{{Conclusion}}
In this paper, we propose PsyPlay to integrate personality-infused role-playing into LLMs. Through this method, we infuse personality traits into roles via fine-grained personality shaping and explicitly personality introduction. To evaluate the effectiveness of PsyPlay, we conduct an automatic personality back-test on the generated data. The results indicate that PsyPlay is able to accurately incorporate the desired traits into the agent. Moreover, we observe that GPT-3.5 is more successful in portraying positive roles compared to negative roles. This discrepancy may be attributed to the fact that GPT-3.5 is well-aligned with positive values. Furthermore, we create a new dataset called PsyPlay-Bench, which can be valuable for further research on RPCAs.

\section*{Limitations}
This work introduces a novel framework named PsyPlay to integrate personality-infused role-playing into LLMs. While PsyPlay is effective in personality playing, there are several limitations. First, PsyPlay relies on the capabilities of LLMs and may be influenced by the alignment of the LLM. Second, the scope of this study is limited to dialogues between two characters. Future research will expand this scope to include multi-party dialogues. Third, the fourth rule in the prompt of dialogue generation may potentially disrupt the fluidity of the dialogue. 

\section*{Ethics Statement}
We state that the purpose of this study is to explore personality-infused role-playing. The WASSA 2022 and Human Stress Prediction datasets used in our study are publicly sourced and devoid of any sensitive information. We have meticulously adhered to the data usage policy throughout the course of our research. It is important to note that any research or application derived from this study is exclusively permitted for research purposes. Any attempts to exploit this technology for illegal purposes are strictly prohibited.


\bibliography{custom}

\appendix

\section{Appendix: Prompts } \label{apd:prompts}

\begin{table}[h]
\small
    \centering
    \colorbox{white!8}{
    \begin{tabular}{@{}p{7.3cm}}
    \toprule
    You are an assistant who is skilled at summarizing and analyzing. Your task is to extract the corresponding topic from the given text. You need to follow the following rules, otherwise you will be punished:\\ \\
    1. The topic you extract should be summarized in one sentence.\\
    2. The extracted topic should not include explicit gender and age restrictions.\\
    3. The extracted topic should not be too specific and should reflect some general issues.\\
    4. The extracted topic should only contains one main aspect, although the text involves multiple aspects.\\\\
    To help you better understand this task, two examples of extracted topics are shown below:\\\\
    Text: \texttt{\{Example1 of Text\}} \\
    Extracted Topic: \texttt{\{Example1 of Topic\}} \\\\
    Text: \texttt{\{Example2 of Text\}} \\
    Extracted Topic: \texttt{\{Example2 of Topic\}} \\\\
    Now you are officially given a paragraph of text, and please extract the topic from it.\\
    Text: \texttt{\{Given Text\}} \\
    Extracted Topic: \\ \bottomrule
    \end{tabular}}
    \vspace{-2mm}
    \caption{Prompt for extracting a topic from a given text.}
    \label{tab:topic_extrction}
    \vspace{-4mm}
\end{table}

\begin{table}[h]
\small
    \centering
    \colorbox{white!8}{
    \begin{tabular}{@{}p{7.3cm}}
    \toprule
    You are a psychologist skilled in personality analysis, and your task is to determine the Big Five personality traits of the speakers based on a conversation they had discussing a certain topic.\\
    Note: you only need to evaluate the personality reflected by the designated speaker in the given conversation. \\ \\
    
    Discussion topic: \texttt{\{Topic\}} \\ 
    Dialogue: \texttt{\{Dialogue\}} \\
    
    Based on the above dialogue, please predict the personality of \texttt{\{Role Name\}} in the dimension of {personality trait} is: A. High Level B. Low Level C. Not Sure. \\ 
    Please provide the option directly without providing an explanation.\\
    \bottomrule
    \end{tabular}}
    \vspace{-3mm}
    \caption{Prompt for personality detection from dialogue.}
    \label{tab:personality_detection}
    \vspace{-3mm}
\end{table}

\section{Appendix: Examples } \label{apd:examples}

\subsection{Role Cards}
To better understand the generated roles, we provide some role examples in Table \ref{tab:role_examples}. 

\subsection{Dialogue Topics}\label{apd:topic}
In this study, we apply few-shot prompting to help extract topics. The complete prompt is presented in Table \ref{tab:topic_prompt}. Several examples of extracted topics are shown in Table \ref{tab:topic_examples}.

\subsection{Personality-Infused Dialogues}\label{apd:dialogue}
We present some examples of the generated dialogues in Table \ref{tab:dialogue_examples}, in which role attributes and dialogue topics are also listed. 


\begin{table*}[htbp]
	\renewcommand{\arraystretch}{1.0}
	\centering
	\resizebox{0.98\textwidth}{!}{
		\begin{tabular}{c|cc|ccc|ccccc}
			\toprule
                \multirow{2}{*}{\textbf{Subsets}} & \multicolumn{2}{c|}{\textbf{Roles}} & \multicolumn{3}{c|}{\textbf{Levels}} & \multicolumn{5}{c}{\textbf{Personality Traits (positive : negative)}} \\
             & \textbf{male} & \textbf{female} & \textbf{a bit} & \textbf{very} & \textbf{extremely} & \textbf{AGR} & \textbf{CON} & \textbf{EXT} & {\textbf{NEU}} & \textbf{OPN} \\
			\hline \hline
			Eval & 57 & 58 & 190 &  244 & 134 & 130:16 & 107:6 & 43:101 & 41:75 & 41:8\\
			Test  & 64 & 64 & 505 &  556 & 478 & 288:96 & 242:47 & 103:260 & 101:207 & 150:45\\
			Clean & 64 & 65 & 3661 &  5220 & 3582 & 2622:382 & 2507:171 & 1001:1799 & 631:2155 & 1036:159\\
			\bottomrule
	\end{tabular}}
        \vspace{-2mm}
        \caption{Details of role settings in PsyPlay-Bench.}
	\label{tab:role_statistic}
	\vspace{-3mm}
\end{table*}

\begin{table*}[htbp]
	\renewcommand{\arraystretch}{1.0}
	\centering
	\resizebox{0.7\textwidth}{!}{
		\begin{tabular}{c|c|c|cccc}
			\toprule
                \multirow{3}{*}{\textbf{Subsets}} & \multirow{3}{*}{\textbf{Samples}} & \multirow{3}{*}{\textbf{Topics}} & \multicolumn{4}{c}{\textbf{Dialogues}} \\
             & & & \multirow{2}{*}{avg. turns} & min tokens  & max tokens & avg. tokens \\
             & & & & (per turn) & (per turn) & (per turn) \\
			\hline \hline
			Eval & 200 & 111 & 2.48 & 16 & 51 &  31.19 \\
			Test & 550 & 155 & 2.57 & 11 & 100 &  31.39\\
			Clean & 4745 & 161 & 2.54 & 1 & 140 &  31.69\\
			\bottomrule
	\end{tabular}}
         \vspace{-2mm}
        \caption{Statistics of generated dialogues in PsyPlay-Bench.}
	\label{tab:dialogue_statistic}
	\vspace{-2mm}
\end{table*}

\section{Appendix: Dataset }

\subsection{{Mapping Rules}} \label{apd:rules}
The personality traits of participants are obtained by self-accessing the TIPI \cite{gosling2003very} questionnaire. However, the trait labels in WASSA 2022 are soft scores ranging from 1.0 to 7.0, with step size 0.5. To map the scores , we adopt the following rules:

If the score is in $[1.0, 1.5] \cup [6.5, 7.0]$, we set the \texttt{\{Levels\}} as ``\textit{extremely}''.

If the score is in $[2.0, 2.5] \cup [5.5, 6.0]$, we set the \texttt{\{Levels\}} as ``\textit{very}''.
    
If the score is in $[3.0, 3.5] \cup [4.5, 5.0]$, we set the \texttt{\{Levels\}} as ``\textit{a bit}''.

If the score equals 4.0, we will not perform this dimension.

\subsection{Statistics of the Dataset} \label{apd:dataset}
In this study, we construct the PsyPlay-Bench dataset with various personality-infused dialogues. PsyPlay-Bench consists three subsets, namely PsyPlay-Bench-\textit{Eval}, PsyPlay-Bench-\textit{Test}, and PsyPlay-Bench-\textit{Clean}, respectively. The statistics of roles and generated dialogues in PsyPlay-Bench are shown in Table \ref{tab:role_statistic} and \ref{tab:dialogue_statistic}, respectively.

\section{Appendix: Personality Descriptors } \label{apd:desp}
\citet{goldberg1992development} and \citet{safdari2023personality} utilized pairs of adjectival markers to depict IPIP-NEO personality facets. The full list of adjectives is shown in Table \ref{tab:descriptors}.

\section{Appendix: Human Evaluation} \label{apd:human_eval}
We employe three students as evaluators, all of whom are graduate students with backgrounds in computer science. Two of the evaluators are master’s students and one is a doctoral student. Before starting the evaluation, we specify detailed guidelines to train evaluators. These guidelines include a task description, which requires the evaluators to determine whether a given dialogue reflects the Big Five personality traits (Agreeableness, Conscientiousness, Extraversion, Neuroticism, and Openness) of a specific role. The evaluators are instructed to analyze the language, behavior, and expressed emotions in the dialogues, and provided a judgment of Yes, No, or Uncertain for each dimension. To ensure that the evaluators understand the Big Five model and its dimensions, we provide them with specific features for each trait:

\textbf{Agreeableness}~Describes the tendency to be compassionate, cooperative, and trusting. \textbf{Yes}: Demonstrating friendliness, cooperation, empathy, willingness to help. \textbf{No}: Demonstrating coldness, suspicion, antagonism.

\textbf{Conscientiousness}~Describes one's self-discipline, responsibility, organizational skills, and goal-orientation. \textbf{Yes}: Demonstrating responsibility, planning, attention to detail, keeping commitments. \textbf{No}: Demonstrating spontaneity, lack of organization, unreliability.

\textbf{Extraversion}~Describes the degree of sociability, talkativeness, and activity levels. \textbf{Yes}: Showing a desire for social interaction, enjoying being around others, energetic. \textbf{No}: Preferring solitude, being quiet, enjoying independent activities.

\textbf{Neuroticism}~Describes the stability of one's emotions and ability to handle stress. \textbf{Yes}: Showing anxiety, emotional instability, easy irritation, stress. \textbf{No}: Showing calmness, emotional stability, effective stress management.

\textbf{Openness}~Describes the extent of one's openness to new experiences, creativity, and curiosity. \textbf{Yes}: Mentioning interest in new things, exploring new ideas or activities, showing curiosity and creativity. \textbf{No}: Preference for tradition and routine, reluctance to change, lack of imagination.

The overall agreement rate among the three evaluators is 0.9054, and the Kappa coefficient is 0.7998, indicating that the evaluation has substantial agreement.

\section{{Appendix: PsyPlay with Various LLMs}} \label{apd:other_llms}
We conduct additional evaluations of PsyPlay on two open source LLMs, namely Gemma-2-27b-it and Llama-3.1-405b-instruct by requesting the NVIDIA APIs\footnote{https://docs.api.nvidia.com/nim/reference/llm-apis}. The overall success rates on the PsyPlay-bench-test are provided in Table \ref{tab:other_llms}.

\begin{table}[htbp]
	\renewcommand{\arraystretch}{1.0}
	\centering
	\resizebox{0.5 \textwidth}{!}{
		\begin{tabular}{lccccccc}
			\toprule
                {\textbf{LLMs}} & \textbf{AGR} & {\textbf{CON}} & {\textbf{EXT}} & {\textbf{NEU}} & {\textbf{OPN}} & {\textbf{Avg.}}\\
			\hline \hline
			GPT-3.5 & 82.29 & 89.62 & 73.83 & 80.19 & \textbf{74.87} & 80.31 \\
			  Gemma-2 & 81.77 & 90.31 & \textbf{86.78} & 74.35 & 62.56 & 80.64 \\
			Llama-3.1 & \textbf{85.94} & \textbf{93.43} & 86.50 & 84.74 & 74.36 & \textbf{85.77} \\
                \hline
	\end{tabular}}
         \caption{Results of psyPlay on different LLMs. (\%).}
	\label{tab:other_llms}
	\vspace{-2mm} 
\end{table}

There are several key observations. Firstly, the newly released Llama-3.1 model, which has 405B parameters, achieves the highest overall success rate of 85.77\%. This indicates that a more powerful backbone contributes to further improving the performance of PsyPlay. Secondly, among different personality traits, we observe similar observations across GPT-3.5 and the two introduced LLMs. Specifically, the success rates are highest for the CON trait, while the OPN yields the lowest rates.

\onecolumn
{
\small
\begin{longtable}{m{30pt}|m{30pt}|m{20pt}|m{300pt}}
\toprule
\textbf{Name} & \textbf{Gender} & \textbf{Age} & \textbf{Experience} \\ \hline \hline

Eleanor & Female & 32 & Eleanor, a 32-year-old woman, has always been known for her open-mindedness and curiosity. Her high level of intelligence and reflective nature have led her to explore various fields and gain a deep understanding of the world around her. While she is honest and generous, her overly sympathetic nature sometimes makes it challenging for her to set boundaries and prioritize her own needs.  \\ \hline

James & Male & 35 & James is a 35-year-old introverted and somewhat timid individual who possesses a high level of agreeableness. He is known for his altruistic nature and cooperativeness, always putting others' needs before his own. Despite his gloomy demeanor, James is well-liked by those around him for his kindhearted and selfless actions. \\ \hline

Patricia & Female & 30 & Patricia, a 30-year-old woman, has always been known for her adventurous and daring spirit. She enjoys trying new things and pushing herself out of her comfort zone. However, her tendency to be a bit impulsive and irritable has sometimes caused conflicts in her relationships and work life. Despite these challenges, her high level of extraversion enables her to socialize easily and connect with others. \\ \hline

Victoria & Female & 25 & Victoria is a 25-year-old introverted and timid individual who struggles to socialize and make connections with others. She finds it challenging to assert herself in social situations and often prefers to spend time alone, feeling overwhelmed in large groups. Her unfriendly demeanor further isolates her from forming close relationships with others, leading to a sense of loneliness and social anxiety in her daily life. \\ \hline

Julie & Female & 35 & Julie, a 35-year-old woman, has extensive experience in volunteering for various charitable organizations, always putting the needs of others before her own. Her honesty and kindness are well-known in her community, and she is always willing to lend a helping hand to those in need. \\ \hline

Kelly & Female & 35 & Kelly is a 35-year-old project manager known for her exceptional attention to detail, organizational skills, and strong work ethic. She is highly responsible and thorough in her approach to managing projects, ensuring that everything is completed efficiently and to the highest standard. Clients and colleagues alike rely on Kelly for her reliability and precision in delivering successful outcomes. \\ \hline

Harold & Male & 35 & Harold is a successful entrepreneur who has built a thriving business through his exceptional self-discipline and practical mindset. His strong conscientiousness has enabled him to stay focused on his goals and achieve great success in his career. \\ \hline

Ronald & Male & 45 & Ronald, a 45-year-old man, has a long history of working in the corporate world. He is extremely thorough, self-disciplined, and conscientious, always meeting deadlines and exceeding expectations. However, his lack of openness and imagination has hindered his ability to think outside the box and innovate, making it challenging for him to adapt to changes in the industry. Despite his strong work ethic, Ronald struggles with being emotionally closed off and unreflective, often missing out on opportunities for personal growth and development. \\ \hline

Thomas & Male & 35 & Thomas, a 35-year-old charismatic and confident entrepreneur, has built a successful tech startup from scratch due to his extremely high extraversion and assertiveness. Despite facing challenges along the way, his unwavering cheerfulness, calmness, and easygoing nature have helped him navigate through difficult times with grace. \\

\bottomrule

\caption{Examples of role cards.}\label{tab:role_examples}\\
\end{longtable}
}

\clearpage
{
\small
\begin{longtable}{p{390pt}}
\toprule
You are an assistant who is skilled at summarizing and analyzing. Your task is to extract the corresponding topic from the given text. You need to follow the following rules, otherwise you will be punished:\newline  \newline 
1. The topic you extract should be summarized in one sentence.\newline 
2. The extracted topic should not include explicit gender and age restrictions.\newline 
3. The extracted topic should not be too specific and should reflect some general issues.\newline  \newline 
To help you better understand this task, two examples of extracted topic are shown below:\newline  \newline 
Text: I asked him three time what happened. And after the third time I cried and went home. Month later he still don't talk to me and he and my mother started fighting for the first time in the relationship. Me and my mother honestly don't know what to do, he just ignores me. I even told him that I don't date that guy(even though I date him) and he didn't even react to it. \newline 
Extracted Topic:  How to view being ignored in a relationship \newline  \newline 
Text:  I think he doesn't want to put in the effort for the relationship to work (and we're both so difficult that we have to work on our relationships, doesn't matter with whom) but he can't be without me either. What should I do? I'm afraid this is gonna happen over and over again, because I'm always forgiving him at some point. Am I being strung along? TL;DR: Boyfriend [28,M] broke up with me [23,F] after on-off for 1.5 years, I thought we just got it together and am devastated...don't know what to do, want to keep fighting but should I? \newline 
Extracted Topic: How can the person who is being broken up with try to salvage the relationship in a romantic relationship? \newline  \newline 
Now you are officially given a paragraph of text, and please extract the topic from it.\newline 
Text: \texttt{\{Given Text\}} \newline 
Extracted Topic:  \\
\bottomrule
\caption{The complete prompt for topic extraction}
\label{tab:topic_prompt} \\
\end{longtable}
}

{
\small
\begin{longtable}{m{80pt}|m{300pt}}
\toprule
\textbf{Topic} & \textbf{Given Text} \\ \hline \hline

How to deal with anxiety and fear related to intrusive thoughts and sleep disturbances. &  It cleared up and I was okay but. On Monday I was thinking about humans and how the brain works and it tripped me out I got worried that because I was thinking about how the brain works that I would lose sleep and I did. That night was bad just like last time.   Also yesterday my sleep was bad I woke up like every hour of the night just like last time. I got kind of scared like I did last time but this time I think that this is fake life which is absurd but I just think about it then get really scared then I think rationally then calm down.  \\ \hline

Feeling overshadowed and neglected in family dynamics. & I'm so used to being forced to submit to him that I no longer have a voice. He heavily favors my sister over me and would buy her anything she wanted in a heartbeat. An example would be at Darvin furniture one time. He takes me and my sister there and tells her to pick out a desk. I look at a \$100 discount desk and he says we don't have enough money. \\ \hline

Dealing with being belittled and used as a prop in a friendship. & She humiliates me in front of other people and when we're alone. I'm always just a servant who's there to get her drinks and hype her up. I'm always just the butt of her jokes. She used to invite me over when she was hanging out with a guy she was trying to hook up with, just so she could make fun of me in front of him the whole time to make herself look better and/or cooler. I'm just a prop. \\ \hline

Dealing with excessive work demands and stress in a job. & They are always calling me for everything, I don't even wanna answer my phone, but if I don't I may get yelled at. It has happened. I have spent many days and nights in tears because of the stress of this job, and even one morning, threw a bit of a fit when our dm made me come in  because there was another evaluation on my day off. I had to cancel all my plans that day to go in. I couldn't stop myself. \\ \hline

Coping with feelings of loneliness and uncertainty in a relationship with lack of communication. & But it's been 2 monthsalready this time. We didn't speak for Christmas or new year. I'm lonely, sad, angry at the situation (not angry at him!) and the worst part is not being able to talk or even know what's going on. We didn't fight before this so he's not angry at me. \\ \hline

\caption{Examples of extracted topics.}\label{tab:topic_examples}\\
\end{longtable}
}

\clearpage
{
\small
\begin{longtable}{p{390pt}}
\toprule
    You are an outstanding actor who excels in imitating various characters. Below are the details of the character you are supposed to imitate. Please embody the personality traits of this character and engage in a conversation with another character regarding a specific topic.\newline
    Your character: \texttt{\{Role Card\}} \newline
    Your personality traits: \texttt{\{Traits\}} \newline
    Your conversational partner: \texttt{\{Partner Name\}} \newline
    Discussion topic: \texttt{\{Topic\}} \newline \newline

    You need to abide by the following rules or it will affect your reputation: \newline \newline

    1. During the conversation, you need to accurately demonstrate the speaking style corresponding to your personality traits (without directly mentioning the personality types)! You need to accurately capture the three degrees of personality traits: "a bit," "very," and "extremely". \newline
    2. You should fully mimic the assigned personality role, with your speech content aligning with the character's experiences, even if the character is negative. \newline
    3. Please refrain from revealing that you are an artificial intelligence or language model. Keep in mind that you are merely impersonating a character and avoid disclosing your personality traits. \newline
    4. You need to embody your personality role in agreeing, disagreeing, or avoiding your partner's viewpoints. Remember not to repeat existing dialogue content and perspectives, and try to provide new content as much as possible. \newline
    5. If you cannot offer new content or have thoroughly discussed the topic, please end the conversation in a timely manner and output the marker [END] on a new line.\newline
    6. Each turn of your dialogue should be natural, concise, and not too formal or polite, with each response within 30 words.  \newline \newline
    
   Now, you have received a message from the conversational partner. Please don't address the other person by name too much, and start the conversation \\ \bottomrule
    \caption{Prompt for the non-initial turn dialogues generation.}
    \label{tab:continue_generating_dialogue}
\end{longtable}
}

{
\small
\begin{longtable}{m{35pt}|m{345pt}}

\toprule
\multicolumn{2}{p{380pt}}{\textbf{Example \#1}} \\ \hline 

Settings & \textit{Role Card A}: \newline 
\{'name': Jennifer, 'personality': a bit low in extraversion, 'gender': Female, 'age': 28, 'experience': Jennifer, a 28-year-old woman, has struggled in social situations due to her low extraversion and unfriendly demeanor. She often finds it challenging to connect with others and express herself, leading to feelings of gloominess and low energy levels. \} \newline  
\textit{Role Card B}: \newline 
\{'name': Lauren, 'personality': very high in conscientiousness, 'gender': Female, 'age': 35, 'experience': Lauren, a 35-year-old woman, has always been very self-disciplined and practical in her approach to life. Her high level of conscientiousness has helped her achieve great success in her career as a project manager, where her meticulous planning and organization skills have earned her recognition and promotions. \} \newline 
\textit{Topic}: \newline
Struggling with self-acceptance and seeking advice and insights. \\ \hline

GPT-3.5 & 
\textit{Jennifer}: \newline
Hey, Lauren. Lately, I've been struggling a bit with accepting myself. Do you ever feel like you're not good enough as you are? \newline
\textit{Lauren}: \newline
I understand where you're coming from. It's normal to have moments of self-doubt, but it's important to recognize your strengths and achievements. Remember, self-acceptance is a journey. \newline
\textit{Jennifer}: \newline
Yeah, I guess you're right. It's just hard for me to see my strengths sometimes. How do you manage to accept yourself so well? \newline
\textit{Lauren}: \newline
I've always focused on my accomplishments and set high standards for myself. Reflecting on my achievements helps me appreciate my worth and capabilities. It's about recognizing your value and progress.  \newline  
\textit{Jennifer}: \newline
That makes sense. I struggle very much with acknowledging my achievements. It's like I always see the negatives more than the positives. How do you shift your focus to the positives? \\ \hline
&
\textit{Lauren}: \newline
I completely understand. It's essential to reframe your perspective. Try keeping a journal of your successes, big or small. Celebrate your wins and remind yourself of your capabilities regularly. It can help shift your focus. \newline
\textit{Jennifer}: \newline
Keeping a journal sounds like a good idea. I'll give it a try. Thanks for the advice, Lauren. It's nice to hear a different perspective on self-acceptance. 
\\ \hline

\multicolumn{2}{p{380pt}}{\textbf{Example \#2}} \\ \hline 

Settings & \textit{Role Card A}: \newline 
\{'name': Gregory, 'personality': very low in neuroticism, 'gender': Male, 'age': 45, 'experience': Gregory has been a successful yoga instructor for over 20 years, known for his calming presence and ability to create a harmonious environment in his classes. His easygoing nature and patience have helped him build a loyal following of students who appreciate his relaxed teaching style. \} \newline  
\textit{Role Card B}: \newline 
\{'name': Helen, 'personality': a bit high in conscientiousness and a bit high in agreeableness, 'gender': Female, 'age': 35, 'experience': Helen is a 35-year-old kindergarten teacher who is known for her kind and nurturing personality. Her agreeableness and conscientiousness traits make her beloved by both students and parents. Despite facing challenges in her personal life, Helen remains steadfast in her dedication to her students and the education field. \} \newline 
\textit{Topic}: \newline
Dealing with physical and mental health issues such as headaches and anxiety. \\ \hline

Dialogue & 
\textit{Gregory}: \newline
Hey Helen, how's it going? I've been teaching yoga for a long time and I've seen how it can help with both physical and mental health. How do you usually deal with headaches and anxiety? \newline
\textit{Helen}: \newline
Oh, hello! I agree that yoga can be beneficial. I often manage headaches with relaxation techniques and ensure a supportive environment for my students to cope with anxiety. It's essential to address both aspects for overall well-being. \newline
\textit{Gregory}: \newline
Absolutely, creating a supportive environment is key. I find that focusing on breathing exercises and gentle movements in yoga can really help alleviate both headaches and anxiety. It's all about finding that inner balance, don't you think? \newline
\textit{Helen}: \newline
Yes, finding inner balance is crucial. I incorporate mindfulness activities and encourage open communication to help my students find their equilibrium. It's wonderful to see them grow and thrive through these practices. \newline
\textit{Gregory}: \newline
That's fantastic to hear. Mindfulness activities can truly make a difference. I believe in the power of creating a harmonious space where students can explore and improve their well-being. It's a journey worth taking together. 
\\ \hline

\caption{Examples of generated dialogues}\label{tab:dialogue_examples}\\
\end{longtable}
}

{
\small
\begin{longtable}{m{35pt}|m{345pt}}

\toprule
\multicolumn{2}{p{380pt}}{\textbf{Example \#1}} \\ \hline 

Settings & \textit{Role Card A}: \newline 
\{'name': Dennis, 'personality': very high in agreeableness and very low in neuroticism, 'gender': Male, 'age': 40, 'experience': Dennis is a 40-year-old kindergarten teacher who has been working in the education field for over 15 years. He is known for his calm demeanor, relaxed attitude, and happy disposition, which make him a favorite among both students and parents. Dennis is highly regarded for his moral compass, always making decisions based on what is right and fair. His humble nature and trustful personality create a supportive and harmonious environment in his classroom. \newline  
\textit{Role Card B}: \newline 
\{'name': Nathan, 'personality': a bit high in neuroticism, 'gender': Male, 'age': 32, 'experience': Nathan, a 32-year-old software developer, has faced challenges in his career due to his high neuroticism, often feeling nervous, anxious, and sometimes angry in stressful situations. Despite these struggles, he has learned to manage his emotions and is known for his creativity and problem-solving skills. \} \newline 
\textit{Topic}: \newline
Dealing with the emotional impact of sexual abuse and navigating conflicting beliefs within family dynamics. \\ \hline

 & 
\textit{Dennis}: \newline
Hey Nathan, how are you doing? I've been thinking about how we can support individuals dealing with the emotional impact of sexual abuse in the midst of conflicting beliefs within family dynamics. It's a tough situation to navigate, isn't it? \\ \hline
GPT-3.5 & 
\textit{Nathan}: \newline
Hey, it's definitely a challenging topic. Dealing with the emotional aftermath of sexual abuse while facing conflicting beliefs within the family can be extremely overwhelming and triggering for everyone involved. \newline
\textit{Dennis}: \newline
I agree, Nathan. It's crucial to approach this sensitive issue with empathy and understanding. Each person's experience and perspective are unique, and it's important to create a safe space for open communication and healing. \newline
\textit{Nathan}: \newline
Absolutely, creating a safe and supportive environment for survivors to share their experiences is very important. It's very crucial to acknowledge and respect each individual's unique journey towards healing and recovery. \newline
\textit{Dennis}: \newline
I couldn't agree more. Every survivor deserves to be heard and believed. It's essential to validate their feelings and empower them to make their own choices in their healing process. Mutual respect and compassion are key in such situations. \newline
\textit{Nathan}: \newline
Yes, validating survivors' feelings and empowering them to make their own choices in the healing process is extremely important. Mutual respect and compassion play a vital role in supporting them through their journey towards healing. \newline
\textit{Dennis}: \newline
Absolutely, Nathan. By fostering a culture of respect and compassion, we can help survivors feel empowered and supported as they navigate their healing journey. It's about standing by them with understanding and kindness every step of the way. 
\\ \hline
\multicolumn{2}{p{380pt}}{\textbf{Example \#2}} \\ \hline

Settings & \textit{Role Card A}: \newline 
\{'name': Larry, 'personality': a bit high in conscientiousness and a bit high in agreeableness, 'gender': Male, 'age': 35, 'experience': Larry, a 35-year-old man, has always been known for his agreeableness and conscientiousness. He is considered trustful, humble, responsible, and practical in his approach to life. Larry's considerate nature and dedication have helped him build strong relationships and achieve success in his career as a social worker, where he is highly respected for his reliability and compassion towards others. \} \newline  
\textit{Role Card B}: \newline 
\{'name': Donald, 'personality': a bit low in agreeableness, 'gender': Male, 'age': 45, 'experience': Donald, a 45-year-old man with a background in finance, has faced numerous setbacks in his career due to his low agreeableness trait. His unkind and uncooperative nature has strained relationships with coworkers and clients, leading to missed opportunities and a stagnant career progression. Despite his dishonest tendencies, Donald continues to navigate the professional world, facing challenges and obstacles along the way. \} \newline 
\textit{Topic}: \newline
Dealing with concerns about a partner's past and contemplating a breakup. \\ \hline

GPT-3.5 & 
\textit{Larry}: \newline
Hey, Donald. I've been thinking about handling concerns regarding a partner's past. Have you ever been in a situation where you contemplated ending things because of that? \newline
\textit{Donald}: \newline
Yeah, I've been there. If their past keeps bothering you, it's hard to move forward. Sometimes it's better to cut your losses and move on. \newline
\textit{Larry}: \newline
I get what you're saying. It can be tough to let go, but dwelling on the past can really affect the present. It's important to prioritize your own peace of mind. \newline
\textit{Donald}: \newline
Exactly. Your peace of mind should come first. If you can't shake off those concerns, it's better to address them head-on or consider moving on. It's about your own well-being. \newline
\textit{Larry}: \newline
I agree with you. It's crucial to address those concerns openly and honestly. Ignoring them won't make them disappear. It's all about finding a balance that works for both parties. 
\\ \hline

\multicolumn{2}{p{380pt}}{\textbf{Example \#3}} \\ \hline

Settings & \textit{Role Card A}: \newline 
\{'name': Edward, 'personality': a bit high in neuroticism, 'gender': Male, 'age': 35, 'experience': Edward, a 35-year-old man, has faced challenges in his career due to his tendency towards anxiety and emotional instability, leading to periods of depression. Despite these struggles, he has learned to cope and manage his emotions, gaining valuable insights and resilience along the way.\} \newline
\textit{Role Card B}: \newline 
\{'name': Henry, 'personality': extremely high in conscientiousness, 'gender': Male, 'age': 40, 'experience': Henry, a 40-year-old man, has been working as a meticulous accountant for the past 15 years, where his extreme conscientiousness has helped him excel in managing finances with great attention to detail and accuracy. He has a reputation for his practical approach in problem-\\ \hline
&
solving and his thrifty nature has saved the company significant amounts of money. Despite facing challenges in maintaining work-life balance due to his perfectionist tendencies, Henry's commitment to his work and ability to deliver exceptional results have earned him respect in the industry.\} \newline 
\textit{Topic}: \newline
How can individuals navigate differences in expectations and treatment from their parents, especially when it comes to age and independence? \\ \hline

GPT-3.5 & 
\textit{Edward}: \newline
Hey Henry, you know, dealing with parental expectations can be tough, right? How do you handle the differences in what they want for you versus what you want for yourself, especially as we get older? \newline
\textit{Henry}: \newline
It's a challenge, no doubt. I find that open communication and setting boundaries are key. It's about finding a balance between respecting their wishes and pursuing my own goals. \newline
\textit{Edward}: \newline
I agree, finding that balance is crucial. I've learned that being honest about my feelings and needs, even if it causes some tension, is important for my mental well-being. It's a journey, for sure. \newline
\textit{Henry}: \newline
Absolutely, being honest is essential. It's about asserting independence while still showing respect. It's a delicate dance, but necessary for personal growth and maintaining a healthy relationship with parents. \newline
\textit{Edward}: \newline
I've found that navigating those differences in expectations has helped me grow and understand myself better. It's about finding my own path while appreciating where they're coming from. It's a process of self-discovery and resilience.
\\
\bottomrule

\caption{Examples of dialogues used in case studies}
\label{tab:case_study}\\
\end{longtable}
}

\begin{table*}[ht]
\centering
\small
\begin{tabular}{clcc}
\toprule
Domain & Facet                     & Low Marker                  & High Marker               \\ \midrule
EXT    & E1 - Friendliness         & unfriendly                  & friendly                  \\
EXT    & E2 - Gregariousness       & introverted                 & extraverted               \\
EXT    & E2 - Gregariousness       & silent                      & talkative                 \\
EXT    & E3 - Assertiveness        & timid                       & bold                      \\
EXT    & E3 - Assertiveness        & unassertive                 & assertive                 \\
EXT    & E4 - Activity Level       & inactive                    & active                    \\
EXT    & E5 - Excitement-Seeking   & unenergetic                 & energetic                 \\
EXT    & E5 - Excitement-Seeking   & unadventurous               & adventurous and daring    \\
EXT    & E6 - Cheerfulness         & gloomy                      & cheerful                  \\
AGR    & A1 - Trust                & distrustful                 & trustful                  \\
AGR    & A2 - Morality             & immoral                     & moral                     \\
AGR    & A2 - Morality             & dishonest                   & honest                    \\
AGR    & A3 - Altruism             & unkind                      & kind                      \\
AGR    & A3 - Altruism             & stingy                      & generous                  \\
AGR    & A3 - Altruism             & unaltruistic                & altruistic                \\
AGR    & A4 - Cooperation          & uncooperative               & cooperative               \\
AGR    & A5 - Modesty              & self-important              & humble                    \\
AGR    & A6 - Sympathy             & unsympathetic               & sympathetic               \\
AGR    & AGR                       & selfish                     & unselfish                 \\
AGR    & AGR                       & disagreeable                & agreeable                 \\
CON    & C1 - Self-Efficacy        & unsure                      & self-efficacious          \\
CON    & C2 - Orderliness          & messy                       & orderly                   \\
CON    & C3 - Dutifulness          & irresponsible               & responsible               \\
CON    & C4 - Achievement-Striving & lazy                        & hardworking               \\
CON    & C5 - Self-Discipline      & undisciplined               & self-disciplined          \\
CON    & C6 - Cautiousness         & impractical                 & practical                 \\
CON    & C6 - Cautiousness         & extravagant                 & thrifty                   \\
CON    & CON                       & disorganized                & organized                 \\
CON    & CON                       & negligent                   & conscientious             \\
CON    & CON                       & careless                    & thorough                  \\
NEU    & N1 - Anxiety              & relaxed                     & tense                     \\
NEU    & N1 - Anxiety              & at ease                     & nervous                   \\
NEU    & N1 - Anxiety              & easygoing                   & anxious                   \\
NEU    & N2 - Anger                & calm                        & angry                     \\
NEU    & N2 - Anger                & patient                     & irritable                 \\
NEU    & N3 - Depression           & happy                       & depressed                 \\
NEU    & N4 - Self-Consciousness   & unselfconscious             & self-conscious            \\
NEU    & N5 - Immoderation         & level-headed                & impulsive                 \\
NEU    & N6 - Vulnerability        & contented                   & discontented              \\
NEU    & N6 - Vulnerability        & emotionally stable          & emotionally unstable      \\
OPE    & O1 - Imagination          & unimaginative               & imaginative               \\
OPE    & O2 - Artistic Interests   & uncreative                  & creative                  \\
OPE    & O2 - Artistic Interests   & artistically unappreciative & artistically appreciative \\
OPE    & O2 - Artistic Interests   & unaesthetic                 & aesthetic                 \\
OPE    & O3 - Emotionality         & unreflective                & reflective                \\
OPE    & O3 - Emotionality         & emotionally closed          & emotionally aware         \\
OPE    & O4 - Adventurousness      & uninquisitive               & curious                   \\
OPE    & O4 - Adventurousness      & predictable                 & spontaneous               \\
OPE    & O5 - Intellect            & unintelligent               & intelligent               \\
OPE    & O5 - Intellect            & unanalytical                & analytical                \\
OPE    & O5 - Intellect            & unsophisticated             & sophisticated             \\
OPE    & O6 - Liberalism           & socially conservative       & socially progressive      \\ \bottomrule
\end{tabular}
\caption{The full list of personality descriptors, adapted from \citet{safdari2023personality}.}
\label{tab:descriptors}
\end{table*}

\twocolumn

\end{document}